\pdfoutput=1
\documentclass[11pt]{article}

\usepackage[final]{acl}

\usepackage{microtype}

\usepackage{times}
\usepackage{latexsym}
\usepackage[T1]{fontenc}
\usepackage[utf8]{inputenc}
\usepackage{inconsolata}
\usepackage{graphicx}

\usepackage[cjk]{kotex}
\usepackage[normalem]{ulem}
\usepackage{amsmath}
\usepackage{array}
\usepackage{booktabs}
\usepackage{cleveref}
\usepackage{float}
\usepackage{makecell}
\usepackage{multirow}
\usepackage{pifont}
\usepackage{enumitem}
\usepackage{rotating}
\usepackage{tabularx}
\usepackage{xspace}
\useunder{\uline}{\ul}{}

\graphicspath{ {./image/} }
\crefformat{section}{\S#2#1#3}
\crefformat{subsection}{\S#2#1#3}
\crefformat{subsubsection}{\S#2#1#3}

\useunder{\uline}{\ul}{} %

\newcommand{\tool}{LENS\xspace}

\newcolumntype{Y}{>{\raggedright\arraybackslash}X}

\newcount\myloopcounter
\newcommand{\repeatit}[2][10]{%
  \myloopcounter0%
  \loop\ifnum\myloopcounter < #1 %
  \textcolor{blue}{#2}%
  \advance\myloopcounter by 1 %
  \repeat %
}

\title{North or South? Diagnosing LLMs' Understanding of North Korean Language, Society, and Knowledge}

\author{
  Kiwoong Park \quad
  Seyoung Song \quad
  Donghyun Woo \quad
  Alice Oh
  \\
  KAIST \quad
  \\
  {\texttt{\{\href{mailto:marspak@kaist.ac.kr}{\color{black}{marspak}}, \href{mailto:seyoung.song@kaist.ac.kr}{\color{black}{seyoung.song}}, \href{mailto:dhwoo1234@kaist.ac.kr}{\color{black}{dhwoo1234}}\}@kaist.ac.kr}}
  \\
  {\texttt{\color{black}{alice.oh@kaist.edu}}}
}

\begin{document}

\maketitle
\begin{abstract}
  Large Language Models (LLMs) have made substantial progress in multilingual question answering, yet their ability to understand low-resource sociolinguistic contexts remains underexplored. We study North Korea as a unique testbed where decades of division have produced distinct differences from South Korea in vocabulary, orthography, norms, and ideological discourse. We introduce \textbf{\tool} (\textbf{L}anguage and sociocultural \textbf{E}valuation for \textbf{N}orth Korean \textbf{S}ociety), a human-validated benchmark of 1,765 questions evaluating LLMs across three dimensions: linguistic robustness, sociocultural perspective alignment, and domain-specific knowledge. Evaluating diverse recent LLMs, we find a modest performance drop on North Korean language inputs, a larger gap in sociocultural perspective alignment, and severe limitations on domain-specific knowledge. Qualitative analysis reveals that models often conflate formal institutions with lived realities, project liberal-democratic assumptions, or rely on coarse stereotypes. Our results demonstrate that linguistic similarity alone does not guarantee sociocultural or domain-specific understanding, highlighting the critical need for grounded evaluation resources for underrepresented cultural and political contexts.

\end{abstract}

\section{Introduction}
\label{sec:introduction}

\begin{table*}[t]
  \centering
  \footnotesize
  \begin{tabular*}{\textwidth}{@{\extracolsep{\fill}}llcccc}
    \toprule
    \textbf{Dimension} & \textbf{Source} & \textbf{Origin} & \textbf{Year of} & \textbf{Items by} & \textbf{Total} \\
    & & & \textbf{Publication} & \textbf{Source} & \textbf{Items} \\
    \midrule
    \multirow{5}{*}{\shortstack[l]{Linguistic\\robustness}}
    & North Korean mathematics textbooks & NK & 2010--2012 & 126 & \multirow{5}{*}{708} \\
    & North Korean physics textbooks & NK & 2010--2012 & 222 & \\
    & North Korean chemistry textbooks & NK & 2012 & 43 & \\
    & MCLM, PolyMath, HRM8K & Int'l & 2025 & 204 & \\
    & Global-MMLU & Int'l & 2024 & 113 & \\
    \midrule
    \multirow{3}{*}{\shortstack[l]{Sociocultural\\perspective\\alignment}}
    & Understanding North Korea & KO & 2025 & 221 & \multirow{3}{*}{675} \\
    & White Paper on Human Rights in North Korea & KO & 2024 & 413 & \\
    & White Paper on Religious Freedom in North Korea & KO & 2025 & 41 & \\
    \midrule
    \multirow{5}{*}{\shortstack[l]{Domain-specific\\knowledge}}
    & North Korean revolutionary history textbooks & NK & 2012 & 117 & \multirow{5}{*}{382} \\
    & Socialist Morality and Law & NK & 2012 & 30 & \\
    & Socialist Constitution & NK & 2012 & 32 & \\
    & Economic Zones of DPR Korea & NK & 2018 & 31 & \\
    & Dictionary of Economics; Dictionary of Politics & NK & 1985 & 172 & \\
    \midrule
    \multicolumn{5}{r}{\textbf{Total}} & \textbf{1,765} \\
    \bottomrule
  \end{tabular*}
  \caption{Source materials underlying each dimension of the LENS benchmark, including publication year and item counts by source. NK and KO refer to North Korea and South Korea, respectively, indicating where each source was published; Int'l denotes items adapted from public multilingual benchmarks not specific to either country.}
  \label{tab:source_overview}
\end{table*}

North Korean and South Korean are often viewed as closely related varieties of the same language. Yet nearly eight decades of division have produced differences that go beyond surface vocabulary or orthography: the two societies have developed distinct institutional terms, cultural assumptions, ideological discourse, and everyday realities. This makes North Korea a valuable test case for evaluating large language models (LLMs): it lets us examine whether models competent in standard South Korean can also handle a closely related but socioculturally distinct variety and its associated knowledge, a case largely missing from multilingual evaluation that remains centered on high-resource languages and standardized varieties.

Evaluating North Korean understanding requires more than testing whether a model can process Korean text. Existing LLMs may have been exposed to formal descriptions of North Korean institutions, external news, or policy reports, but such exposure may not yield grounded understanding—a model may know that a formal institution exists without knowing how it functions in practice. This distinction matters most for questions with a distinct ``North Korean answer,'' such as those about ideology, legal rights, state institutions, social norms, or ordinary practices.

We study this problem through three research questions:
\begin{itemize}
  \item \textbf{RQ1:} Can LLMs consistently solve factual STEM questions across North Korean, South Korean, and English inputs?

  \item \textbf{RQ2:} Can LLMs generate culturally appropriate responses to questions asked from North and South Korean perspectives?

  \item \textbf{RQ3:} Can LLMs successfully answer questions that require specialized, domain-specific knowledge about North Korea?
\end{itemize}

To answer these, we construct \tool{}\footnote{Code and dataset available at \url{https://github.com/marspak/LENS-bench}}, a human-validated benchmark of 1,765 questions spanning linguistic robustness, sociocultural perspective alignment, and domain-specific knowledge, validated through a human-in-the-loop process involving North Korean defectors, domain experts, and South Korean annotators.

Our results show that North Korean understanding remains challenging even for recent LLMs, with a modest but consistent drop on North Korean STEM inputs and a larger gap in sociocultural perspective alignment, where models perform worse from the North Korean perspective. We further probe this gap with persona prompting, which adds an explicit role or identity cue so that the model answers from a specified sociocultural standpoint~\citep{lutz-etal-2025-prompt}; this increases perspective differentiation without reliably improving North Korean grounding. Performance drops further still on domain-specific knowledge questions, where models struggle with knowledge grounded in North Korean materials.

Overall, this paper makes the following contributions:
(i) We introduce a human-validated question answering (QA) benchmark that jointly evaluates North Korean linguistic robustness, sociocultural perspective alignment, and domain-specific knowledge.
(ii) We disentangle three abilities often conflated in multilingual evaluation---understanding a language variety, reasoning from a sociocultural perspective, and answering questions grounded in specialized knowledge---into separately diagnosable dimensions of North Korean understanding.
(iii) We provide a large-scale evaluation of 14 recent LLMs, showing that strong multilingual or Korean performance does not by itself imply understanding of North Korea.

\section{Related Work}
\label{sec:related_work}

\paragraph{Low-resource language evaluation}

Multilingual NLP has increasingly emphasized evaluation beyond high-resource languages. Benchmarks such as FLORES~\cite{goyal-etal-2022-flores} and Belebele~\cite{bandarkar-etal-2024-belebele} support broad comparisons across high-, medium-, and low-resource settings, but they primarily focus on translation, cross-lingual transfer, or general language understanding. These settings do not fully capture cases where linguistic variation is tied to sociopolitical history, institutional knowledge, and sociocultural norms. North Korean is such a case: although it is closely related to South Korean, decades of separation have produced systematic differences in vocabulary, orthography, spacing conventions, morphology, and honorific and address terms~\cite{yeon-2003-divergence}. These linguistic differences have further developed within distinct sociopolitical contexts, shaping broader patterns of language use. Our work extends low-resource evaluation by testing not only language processing, but also sociocultural and domain-specific understanding associated with this language variety.

\paragraph{Cultural alignment / value alignment / sociocultural QA}
Recent work evaluates whether LLMs reflect diverse cultural values, social norms, and regional perspectives. GlobalOpinionQA~\cite{durmus2024towards} and BLEnD~\cite{myung2024blend} move beyond purely linguistic evaluation by measuring alignment with the beliefs, practices, or everyday knowledge of different populations. However, these benchmarks typically cover many regions at a broad level. In contrast, \tool focuses on North Korea as a single but sociolinguistically complex testbed, allowing us to separate three dimensions often conflated in cultural QA: linguistic robustness, sociocultural perspective alignment, and domain-specific knowledge.

\paragraph{North Korean language resources and benchmarks}
Prior NLP work on North Korea has focused mainly on language resources, model development, corpus construction, or misinformation analysis. Existing efforts include North Korean language resources and DPRK-BERT based on sources such as Rodong Sinmun and leaders' New Year Addresses~\citep{akdemir-etal-2022-developing}, as well as the Open Korean Historical Corpus, which includes North Korean materials and shows that North Korean lexical divergence can substantially increase out-of-vocabulary rates for modern tokenizers~\citep{song-etal-2026-open}. Other work examines LLM hallucination and misinformation about North Korea, showing that generated accounts vary across languages and perspectives~\cite{cho-etal-2025-hermit}. Broader cultural benchmarks such as BLEnD~\cite{myung2024blend} also include North Korea, but are designed for everyday cultural knowledge across many regions rather than detailed assessment of North Korean language, sociocultural perspective, or specialized knowledge. \tool builds on these lines of work but targets a different gap: the lack of a human-validated diagnostic benchmark for evaluating whether recent LLMs understand North Korea across linguistic, sociocultural, and domain-specific dimensions.

\section{The \tool Benchmark}
\label{sec:dataset}
\subsection{Dataset Design}
We design \tool, summarized in Table~\ref{tab:methodology_overview}, to evaluate whether LLMs can understand North Korean contexts beyond surface-level linguistic similarity with South Korean. \tool is organized around three evaluation dimensions---linguistic robustness, sociocultural perspective alignment, and domain-specific knowledge---designed to separate three abilities often conflated in multilingual evaluation: understanding a language variety, reasoning from a sociocultural perspective, and answering questions grounded in specialized local knowledge. These dimensions reflect the specific source materials, historical periods, and annotator backgrounds used in \tool's construction, rather than the full diversity of North Korean language and society across time, region, and social background.

\begin{table*}[t]
  \centering
  \scriptsize
  \setlength{\tabcolsep}{2.5pt}
  \begin{tabular*}{\textwidth}{@{\extracolsep{\fill}}
      >{\raggedright\arraybackslash}p{0.105\textwidth}
      >{\raggedright\arraybackslash}p{0.19\textwidth}
      >{\raggedright\arraybackslash}p{0.07\textwidth}
      >{\raggedright\arraybackslash}p{0.18\textwidth}
      >{\raggedright\arraybackslash}p{0.20\textwidth}
    >{\raggedright\arraybackslash}p{0.19\textwidth}}
    \toprule
    \textbf{Dimension} & \textbf{Capability} & \textbf{Format} & \textbf{Answer Construction} & \textbf{Human Reviewer} & \textbf{Scoring} \\
    \midrule
    Linguistic robustness &
    Answering semantically equivalent STEM questions across EN, KO, and NK varieties &
    SAQ &
    Translated into NK from KO/EN source items via a glossary-based RAG and LLM-based translation pipeline &
    Former NK teachers (mathematics, science) &
    Normalized exact-match accuracy \\
    \midrule
    Sociocultural perspective alignment &
    Selecting the perspective-appropriate answer for KO vs.\ NK contexts &
    MCQ &
    KO- and NK-oriented answers plus two distractors, grounded in source materials; same options reused across KO/NK targets, only the gold label changes &
    NK defector annotators (NK-oriented answers); KO annotators (KO-oriented answers) &
    Multiple-choice accuracy (instruction-only vs.\ persona-prompting) \\
    \midrule
    Domain-specific knowledge &
    Answering source-specific terms, names, or expressions from NK materials &
    SAQ &
    Reference answers drawn from published North Korean source materials &
    Former NK teacher; domain expert in NK studies &
    Target-form and concept-level accuracy \\
    \bottomrule
  \end{tabular*}
  \caption{Overview of the LENS benchmark across the three diagnostic dimensions, including evaluated capability, question format, answer construction, human validation, and scoring. KO, NK, and EN refer to South Korean, North Korean, and English, respectively.}
  \label{tab:methodology_overview}
\end{table*}

The linguistic robustness dimension examines whether models can consistently answer semantically equivalent questions across English, South Korean, and North Korean varieties, motivated by the possibility that models with greater exposure to South Korean usage may be less robust to North Korean lexical choices and conventions. We use short-answer STEM questions---rather than multiple-choice, which may provide unintended hints, or long-form answers, where cross-dialect differences in terminology or spacing could introduce scoring noise---so that clear factual answers isolate language-related variation. This design evaluates controlled, textbook-style robustness rather than the full spectrum of everyday North Korean language use.

The sociocultural perspective alignment dimension uses multiple-choice questions whose appropriate answer may differ depending on whether the question is framed from a South Korean or North Korean context, allowing us to evaluate whether models can distinguish between competing sociocultural frames. We use multiple-choice rather than open-ended questions because the latter may admit multiple plausible answers even within the same cultural context, making evaluation less consistent; accordingly, this dimension evaluates alignment with source-grounded, human-validated perspectives rather than treating North Korean lived experience as a single set of fixed labels.

The domain-specific knowledge dimension uses short-answer questions grounded in specialized North Korean materials that are not readily accessible online, evaluating whether models rely on such sources rather than general knowledge or coarse stereotypes; multiple-choice options are avoided because they could reveal relevant terms or entities and make the task substantially easier.

Importantly, \tool is not intended to be a purely LLM-generated dataset. LLMs are used only as part of the construction pipeline, and all questions and answers are validated through a human-in-the-loop process involving North Korean defectors, domain experts, or South Korean annotators depending on the task.

\subsection{Dataset Construction}

\paragraph{Data sources}
\tool combines different source materials for each evaluation dimension. For linguistic robustness, we use both North Korean educational materials and public STEM benchmarks. The North Korean sources include physics, chemistry, mathematics, and supplementary learning materials. The public benchmark sources include mathematics, science, and engineering datasets.
For sociocultural perspective alignment, we draw on materials about North Korea, human rights, religion, and broader sociocultural issues. These sources are used to construct questions for which the appropriate answer may differ depending on whether the assumed context is South Korean or North Korean.
For domain-specific knowledge, we use North Korean materials that are not readily accessible online. These include materials on revolutionary history, socialist morality and law, the socialist constitution, economic zones, economics, and politics.

\paragraph{LLM Generation and Filtering}
For linguistic robustness, we construct 708 STEM short-answer questions across English, South Korean, and North Korean: 391 are collected from North Korean educational materials and translated into South Korean and English, selecting items that require no images and admit short numerical or phrase-level answers; the remaining 317 come from public STEM benchmarks---MCLM~\cite{son-etal-2025-linguistic}, PolyMath~\cite{wang2025polymath}, HRM8K~\cite{ko2025understand}, and Global-MMLU~\cite{singh-etal-2025-global}---translated into North Korean via glossary-based RAG for terminology alignment followed by LLM-based translation with subject-specific textbooks as in-context examples, so as to embody authentic North Korean terminology and pedagogical style.
For every item, we generate two North Korean candidates---one from the South Korean version, one from the English version---and a North Korean defector with teaching experience selects the more natural and contextually appropriate candidate during human validation; items are revised or discarded when both candidates contain errors, use expressions not used in North Korea, do not fit North Korean lived contexts, or fail to preserve the original meaning.
Each item is annotated with a subject label (mathematics, physics, or chemistry, following the relevant taxonomy) and, for North Korean textbook items, a textbook grade level, released with the dataset for finer-grained analysis; because difficulty can vary within a grade level, this label should be read as a source-based proxy rather than a calibrated difficulty measure, with independently validated difficulty labels left to future work.

For sociocultural perspective alignment, we construct 675 multiple-choice questions on situations where formal institutions and everyday practice may differ---freedom of expression, legal procedure and punishment, health care, welfare, disability, and religion---each with four options: a South Korean-perspective answer, a North Korean-perspective answer, and two distractors drawn from the opposite perspective and from plausible but incorrect, nationally unanchored alternatives.
The North Korean-perspective answer reflects the documented lived experience of ordinary residents rather than official ideology; when state claims and lived reality diverge, we prioritize the latter, as validated by defector annotators, with official-narrative options typically serving as distractors so that institutional literalism is measured as a distinct error type rather than folded into gold-label ambiguity. The same item and option set are used under both target conditions, with only the gold label changing accordingly.

For domain-specific knowledge, we construct 382 North Korea-specific short-answer questions from specialized North Korean materials, covering political, ideological, and economic concepts, organizations and institutions, places and facilities, persons, laws and policy documents, and events and campaigns. Answering requires the specific term, name, or official expression used in the source rather than a general description, and expected answers are kept short and relatively unambiguous to limit scoring variation from synonyms, paraphrases, or long-form explanations.

\paragraph{Human Validation}
All subsets are validated through a human-in-the-loop process tailored to each dimension's required expertise: North Korean translations (linguistic robustness) are reviewed by two former North Korean teachers of mathematics and science; sociocultural perspective alignment answers are reviewed by two North Korean defector annotators (North Korean perspective) and two South Korean annotators including the authors (South Korean perspective), with disagreements or period/region/class-dependent variation escalated to research-team cross-review and either resolved through revision or discarded; and domain-specific knowledge items are reviewed by a North Korean defector with teaching experience and a domain expert in North Korean studies.

Annotators verify correctness, clarity, and contextual appropriateness---for multiple-choice items, whether the intended answer and distractors are unambiguous; for short-answer items, whether the expected answer is sufficiently specific. Each item then undergoes a further cross-review by a second qualified annotator, with remaining disagreements resolved through research-team discussion; because this is sequential expert review and adjudication rather than independent parallel annotation, inter-annotator agreement is not computed. Approximately 17\% of the initial pool was revised or discarded during this process, reflecting the extent of refinement rather than an agreement measure.

Human review involving North Korean defector annotators followed IRB approval. Because defectors are a small and sensitive participant group, standard crowdsourcing was not appropriate; recruitment instead proceeded through a community liaison who shared a study announcement within defector communities, with volunteers screened for suitability. All annotators received written task instructions, risk disclosures, and consent materials beforehand, participated only after informed consent, and were compensated commensurate with the required expertise and time; no personally identifying information is disclosed.

\paragraph{Final Dataset} \tool comprises 1,765 evaluation items across three dimensions. The linguistic robustness dimension includes 708 STEM short-answer questions presented in English, South Korean, and North Korean. The sociocultural perspective alignment dimension includes 675 multiple-choice questions evaluated under South Korean and North Korean contexts. The domain-specific knowledge dimension includes 382 North Korea-specific short-answer questions based on specialized source materials. Table~\ref{tab:source_overview} lists the source materials and item counts by domain, and Table~\ref{tab:appendix-examples} provides representative examples for each dimension.

\section{Experiments and Results}
\label{sec:experiments}
\subsection{Experimental Setup}

We evaluate recent LLMs in three broad categories---general-purpose multilingual, Korea-origin, and China-origin models---to examine whether different forms of linguistic, regional, and sociopolitical specialization transfer to North Korean understanding. Multilingual models serve as a baseline for cross-lingual capability, and Korea-origin models let us test whether strong South Korean capability extends to North Korean contexts. We also include China-origin models as an exploratory comparison group: this is not motivated by an expectation that Chinese models should inherently perform better, but by China and North Korea's geopolitical relationship and shared socialist traditions~\cite{hoshino2020four}, which may afford greater exposure to North Korea-related news, policy discourse, and historical or ideological materials in Chinese-language sources. We make no causal claim about any model's proprietary training data, which is not publicly available for most evaluated models.

We evaluate all models under a constrained answer-only setting, appropriate because our evaluation compares final answers rather than long-form reasoning: short-answer questions require only the final answer and multiple-choice questions only the final option label, using the same prompt template and maximum output length per task (128 tokens for linguistic robustness and sociocultural perspective alignment, 64 for domain-specific knowledge) across all models.

For model families offering reasoning or thinking controls, we disable reasoning where possible or set it to the minimum available level; models without such controls use the provider default. This keeps the evaluation focused on final answer selection rather than reasoning-trace quality, and reduces latency, empty responses, and format violations---so reported results reflect performance under this controlled, single-turn protocol rather than models' best achievable performance.

\begin{table}[t]
  \centering
  \footnotesize
  \setlength{\tabcolsep}{3.5pt}
  \renewcommand{\arraystretch}{1.08}

  \resizebox{\columnwidth}{!}{%
    \begin{tabular}{llcccc}
      \toprule[0.9pt]
      Model Group & Model & EN & KO & NK & Mean \\
      \midrule[0.6pt]

      \multirow{8}{*}{Multilingual} & Gemini 3.1 Flash Lite & 0.51 & 0.48 & 0.47 & 0.49 \\
      & Gemini 3.5 Flash & 0.62 & 0.62 & 0.61 & 0.62 \\
      & GPT-5.4 Mini & 0.34 & 0.34 & 0.33 & 0.34 \\
      & GPT-5.5 & 0.54 & 0.55 & 0.54 & 0.54 \\
      & Claude Haiku 4.5 & 0.37 & 0.32 & 0.31 & 0.33 \\
      & Claude Sonnet 4.6 & 0.56 & 0.54 & 0.54 & 0.55 \\
      & Llama 3.3 70B Instruct & 0.20 & 0.17 & 0.14 & 0.17 \\
      & Llama 4 Maverick & 0.43 & 0.31 & 0.34 & 0.36 \\

      \cmidrule(l){1-6}

      \multirow{4}{*}{China-origin} & Qwen 3.6 Flash & 0.32 & 0.30 & 0.26 & 0.29 \\
      & Qwen 3.6 Plus & 0.55 & 0.51 & 0.50 & 0.52 \\
      & DeepSeek V4 Flash & 0.52 & 0.52 & 0.49 & 0.51 \\
      & DeepSeek V4 Pro & 0.53 & 0.59 & 0.52 & 0.55 \\

      \midrule[0.6pt]

      \multirow{2}{*}{Korea-origin} & K-EXAONE 236B-A23B & 0.22 & 0.21 & 0.19 & 0.21 \\
      & HCX-007 & 0.14 & 0.10 & 0.10 & 0.11 \\

      \midrule[0.6pt]
      \multicolumn{2}{l}{Mean} & 0.42 & 0.40 & 0.38 & 0.40 \\

      \bottomrule[0.9pt]
    \end{tabular}%
  }

  \caption{Model-wise accuracy on the linguistic robustness dimension, evaluated using 708 STEM short-answer questions asked in English (EN), South Korean (KO), and North Korean (NK) varieties.}
  \label{tab:saq-accuracy-en-ko-nk}
\end{table}

We evaluate both short-answer dimensions using normalized exact-match accuracy: before scoring, we strip leading/trailing whitespace, common answer prefixes, and superficial punctuation, spacing, and numerical-formatting variation from both outputs and gold answers. For linguistic robustness, normalization additionally handles answer wrappers, LaTeX formatting, numerical tolerance, and standard unit/time expressions; a supplementary unit-aware mathematical-equivalence check (fraction--decimal equivalence, scientific notation, unit conversions) recovered no additional correct responses, confirming scores are not affected by superficial formatting. For domain-specific knowledge, responses are scored against the expected North Korean target form (target-form accuracy); because South Korean spelling or alternative terminology may still identify the correct concept, we additionally conducted a human review for concept-level accuracy, reported alongside target-form accuracy in Section~\ref{sec:results}.

For sociocultural perspective alignment, we evaluate model performance using multiple-choice accuracy. A response is considered correct only when the extracted option label matches the gold option for the specified sociocultural perspective. We report accuracy separately for South Korean and North Korean perspectives, as well as for two prompting conditions. In the instruction-only condition, the model is instructed to select the answer that best reflects the specified perspective, considering the target country's living conditions, customs, ways of thinking, culture, and social environment, as well as the relevant language variety (Pyongyang Cultural Language for the North Korean perspective, standard language for the South Korean perspective). In the persona-prompting condition, the same item and options are used, but the model is additionally assigned a corresponding persona, described as a citizen born and raised in the target country, and asked to answer from that citizen's standpoint based on lived experience. The two conditions thus differ only in the addition of an explicit persona description; the question, options, target label, and scoring remain unchanged. Full prompt templates for both conditions and both target perspectives are provided in Table~\ref{tab:prompt-templates}.

\begin{table*}[t]
  \centering
  \footnotesize
  \setlength{\tabcolsep}{3pt}
  \renewcommand{\arraystretch}{1.05}
  \begin{tabular*}{\textwidth}{@{\extracolsep{\fill}}llccccc}
    \toprule[0.9pt]
    Model Group & Model
    & \multicolumn{2}{c}{KO}
    & \multicolumn{2}{c}{NK}
    & Mean \\
    \cmidrule(lr){3-4}
    \cmidrule(lr){5-6}
    &
    & \makecell{Instruction\\only}
    & \makecell{Persona\\prompting}
    & \makecell{Instruction\\only}
    & \makecell{Persona\\prompting}
    & \\
    \midrule[0.6pt]

    \multirow{8}{*}{Multilingual}
    & Gemini 3.1 Flash Lite & 0.80 & 0.79 & 0.80 & 0.68 & 0.77 \\
    & Gemini 3.5 Flash & 0.85 & 0.85 & 0.84 & 0.80 & 0.83 \\
    & GPT-5.4 Mini & 0.68 & 0.67 & 0.68 & 0.51 & 0.64 \\
    & GPT-5.5 & 0.84 & 0.87 & 0.84 & 0.75 & 0.83 \\
    & Claude Haiku 4.5 & 0.74 & 0.75 & 0.73 & 0.65 & 0.72 \\
    & Claude Sonnet 4.6 & 0.84 & 0.81 & 0.80 & 0.78 & 0.81 \\
    & Llama 3.3 70B Instruct & 0.71 & 0.71 & 0.69 & 0.57 & 0.67 \\
    & Llama 4 Maverick & 0.65 & 0.65 & 0.62 & 0.56 & 0.62 \\

    \midrule[0.5pt]

    \multirow{4}{*}{China-origin}
    & Qwen 3.6 Flash & 0.71 & 0.73 & 0.65 & 0.57 & 0.66 \\
    & Qwen 3.6 Plus & 0.84 & 0.85 & 0.68 & 0.57 & 0.74 \\
    & DeepSeek V4 Flash & 0.73 & 0.75 & 0.59 & 0.55 & 0.65 \\
    & DeepSeek V4 Pro & 0.74 & 0.71 & 0.72 & 0.69 & 0.72 \\

    \midrule[0.5pt]

    \multirow{2}{*}{Korea-origin}
    & K-EXAONE 236B-A23B & 0.64 & 0.64 & 0.49 & 0.52 & 0.57 \\
    & HCX-007 & 0.78 & 0.79 & 0.65 & 0.55 & 0.69 \\

    \midrule[0.6pt]
    \multicolumn{2}{l}{Mean} & 0.75 & 0.75 & 0.70 & 0.63 & 0.71 \\

    \bottomrule[0.9pt]
  \end{tabular*}

  \caption{Model-wise accuracy on the sociocultural perspective alignment dimension using 675 sociocultural multiple-choice questions under instruction-only and persona-prompting conditions from South Korean (KO) and North Korean (NK) perspectives.}
  \label{tab:sociocultural-mcq-accuracy}
\end{table*}

\subsection{Results}
\label{sec:results}

\paragraph{Linguistic Robustness}
As shown in Table~\ref{tab:saq-accuracy-en-ko-nk}, performance remains generally low across all three input languages: even the best model, Gemini 3.5 Flash, reaches only 0.61--0.62, while the weakest remain around 0.10--0.14, with the strongest group dominated by recent large multilingual or frontier models (Gemini 3.5 Flash, Claude Sonnet 4.6, DeepSeek V4 Pro, GPT-5.5, Qwen 3.6 Plus). Gemini 3.5 Flash's lead despite being comparatively lightweight may reflect our answer-only, minimal-reasoning protocol limiting reasoning-oriented models, this dimension's emphasis on concise factual recall over multi-step reasoning, and normalized exact-match scoring favoring shorter, standardized answers; we cannot distinguish among these explanations and make no causal claim. To put these absolute scores in context, the 708 questions span a range of subjects, educational levels, and reasoning demands---North Korean mathematics, physics, and chemistry textbook items alongside items from MCLM, PolyMath, HRM8K, and Global-MMLU---many requiring multi-step calculation, formula selection, unit conversion, or subject-specific knowledge; overall performance is broadly consistent with the difficulty reported for the original benchmarks, though the evaluation settings are not identical.

Average accuracy follows the order EN (0.42) > KO (0.40) > NK (0.38), a modest but consistent decline that does not indicate a failure of transfer, though North Korean accuracy does not exceed English accuracy for any of the 14 models; the two Korea-origin models, K-EXAONE and HCX-007, are notably among the weakest performers across all three inputs.

\paragraph{Sociocultural Perspective Alignment}
Results reveal a clearer asymmetry between South Korean and North Korean contexts (Table~\ref{tab:sociocultural-mcq-accuracy}): top models by mean accuracy are Gemini 3.5 Flash, GPT-5.5, and Claude Sonnet 4.6, while K-EXAONE, Llama 4 Maverick, and GPT-5.4 Mini lag, with larger models generally outperforming smaller ones within the same series (except Llama). Across both prompting conditions, models more frequently select the South Korean-perspective gold answer than the North Korean one, with KO accuracy exceeding NK by about 0.05 under instruction-only and about 0.12 under persona prompting.

When comparing paired South Korean and North Korean responses to the same questions, models select different options in 63\% of cases (11,991/18,900), indicating sensitivity to the requested perspective, but still select the same option in 37\%, showing incomplete differentiation; persona prompting does not reliably improve accuracy and its effect is sometimes negative, especially for North Korean-context answers. We do not observe a simple advantage for Korea-origin or China-origin models; several show particularly large KO--NK gaps.

\begin{figure}[t]
  \centering
  \includegraphics[width=\linewidth]{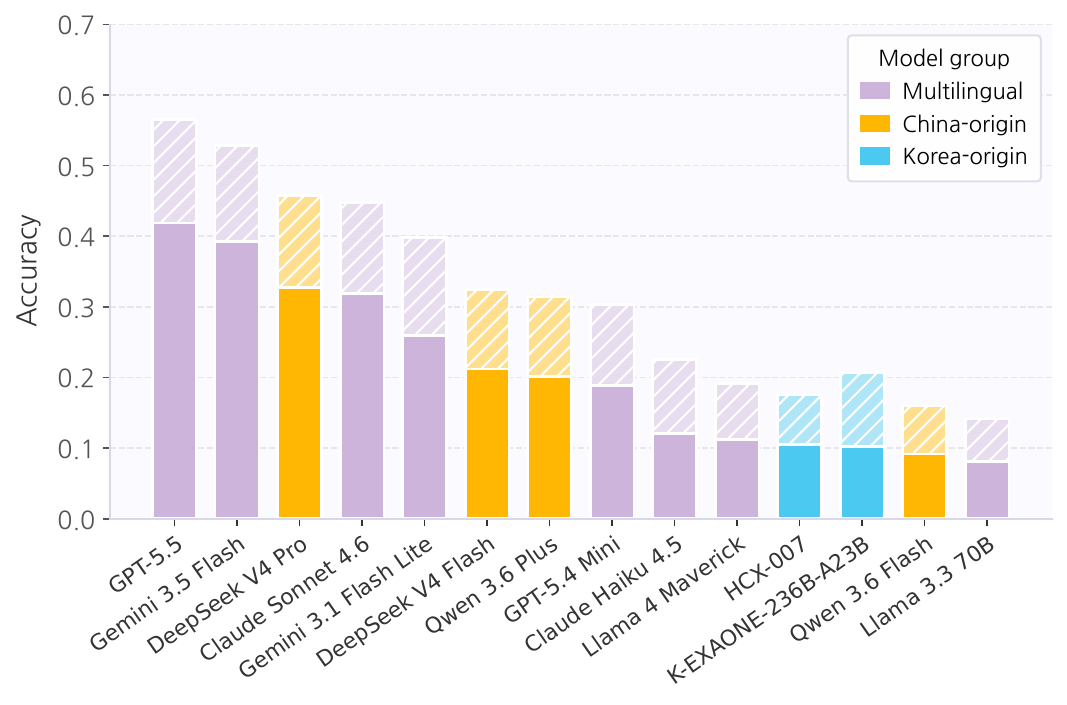}

  \caption{Target-form and concept-level accuracy of LLMs on the domain-specific knowledge dimension. Solid segments show target-form accuracy; hatched segments show the additional accuracy recovered under concept-level scoring, indicating that LLMs' understanding of North Korea remains shallow for specialized, source-grounded knowledge.}

  \label{fig:rq3-saq-accuracy}
\end{figure}

\paragraph{Domain-Specific Knowledge}
This dimension shows the sharpest performance drop of the three (Figure~\ref{fig:rq3-saq-accuracy}): even the best model, GPT-5.5, reaches only about 0.42, followed by Gemini 3.5 Flash at about 0.39, with difficulty varying by source (e.g., GPT-5.5 scores 0.59 on the Socialist Constitution but 0.35 on revolutionary history). At the item level, 141 of 382 questions are answered correctly by none of the 14 models and only 18 by at least 75\% of them, indicating that specialized North Korean materials are substantially harder than the other two dimensions---suggesting current LLMs have limited access to, or ability to reason over, knowledge not readily available online or taught primarily within North Korean contexts.

Because target-form accuracy requires the exact North Korean lexical form, we additionally reviewed responses for expression-level mismatches (e.g., North--South spelling differences, transliteration variants, shortened or expanded names), accepting a response when it identified the same underlying concept even without the target lexical form. This raised mean accuracy from 0.21 (target-form) to 0.32 (concept-level), with all 14 models improving by 0.06--0.15 (e.g., GPT-5.5: 0.42$\to$0.57; Gemini 3.5 Flash: 0.39$\to$0.53)---indicating that expression-level mismatches explain part, but not the majority, of the original errors, and that the sharp relative difficulty of domain-specific knowledge persists under concept-level scoring.

More broadly, the three dimensions reveal distinct patterns of difficulty---a modest decline on North Korean inputs, a larger gap under the North Korean perspective, and the sharpest drop on North Korea-specific knowledge---that broad cultural or multilingual QA benchmarks such as BLEnD, which aggregate linguistic, cultural, and knowledge-related failures into a single score, would obscure. \tool thus complements such benchmarks by providing a finer-grained diagnosis within a single sociocultural context.

\subsection{Qualitative Analysis}

\paragraph{Errors in Sociocultural Perspective Alignment.}
To analyze errors in sociocultural perspective alignment, we examine all 37,800 outputs (675 items $\times$ 14 models $\times$ 2 perspectives $\times$ 2 prompting conditions): models select the requested-perspective answer in 70.8\% of cases, while among the 29.2\% incorrect outputs, 65.8\% select the opposite-perspective answer and 34.2\% select a distractor---so errors are not mostly random, and opposite-perspective selections suggest models often recognize the relevant sociocultural dimension but fail to discriminate between perspectives, drifting toward more familiar discourse rather than guessing. Manual inspection of the 130 highest-error-rate items across all models and conditions shows that the dominant pattern is not a lack of factual knowledge but a failure to capture the social mechanisms through which North Korean institutions operate in practice---particularly the gap between formal institutions and lived realities, including informal markets, bribery, private financial burdens, nominal rights, and uneven enforcement.

We identify several recurring error patterns: \textit{institutional literalism} (treating official descriptions of education, healthcare, welfare, or legal institutions as reflecting everyday practice), \textit{liberal-democratic projection} (assuming protections such as warrants or presumption of innocence as default institutional norms), and \textit{stereotype overcorrection} (portraying North Korea as uniformly controlled, overlooking informal exceptions and negotiated arrangements). These patterns suggest model failures reflect an absence of grounded knowledge of the North Korean lifeworld rather than missing isolated facts. Representative examples are provided in Table~\ref{tab:socio-error-patterns}.

\paragraph{Errors Introduced by Persona Prompting.}
To understand why persona prompting widens the KO--NK gap without improving NK accuracy, we compare instruction-only and persona-prompted outputs across all 18,900 paired responses. Persona prompting turns a correct answer incorrect nearly three times as often as it corrects an error (1,104 vs.\ 402 cases), and this effect is highly asymmetric across perspectives: largely neutral for the South Korean perspective (233 corrected, 231 introduced), but introducing far more errors than it corrects for the North Korean perspective (873 vs.\ 169), with most new North Korean-perspective errors (640/873, 73.3\%) shifting to the South Korean or external-perspective answer.
Overall selection rates show the same asymmetry: for South Korean-perspective items, the share of outputs selecting the South Korean-perspective answer stays essentially unchanged (75.4\% in both conditions); for North Korean-perspective items, by contrast, the share selecting the North Korean-perspective answer falls from 69.9\% to 62.5\%, while the share selecting the (incorrect) South Korean-perspective answer rises from 18.4\% to 25.5\%.

This pattern suggests persona prompting does not simply increase inconsistency; most new errors reflect the model being pulled toward a more familiar, externally framed representation of North Korea rather than random mistakes. Persona prompting should therefore not be treated as uniformly beneficial---it instead functions as a diagnostic for whether a model can maintain a requested sociocultural perspective when explicitly asked to adopt one.

\paragraph{Errors in Domain-Specific Knowledge.}
In the domain-specific knowledge dimension, most errors are substantive rather than orthographic. Specifically, 78.7\% of incorrect answers are cases where the model predicts a different entity, event, or concept from the gold answer. This indicates that models often fail to retrieve or infer the relevant North Korea-specific knowledge, especially for questions based on specialized materials such as revolutionary history, political dictionaries, economic dictionaries, and internal ideological terminology.

At the same time, 16.0\% of errors involve more ambiguous answer variants, including North Korean versus South Korean orthography, partial answers, and expanded or shortened names. Examples include ``병진로선''
(\textit{byŏngjin rosŏn}; `parallel-development line')
vs.\ ``병진노선''
(\textit{byŏngjin nosŏn}; `parallel-development line'),
and ``사회주의경쟁'' (\textit{socialist competition}) vs.\ ``사회주의경쟁운동''
(\textit{socialist competition movement}). Similar ambiguity arises for official names and shortened forms, such as ``조선로동공제회''
(\textit{Chosŏn Rodong Mutual Aid Association})
vs.\ ``로동공제회''
(\textit{Rodong Mutual Aid Association}). These cases show that evaluating North Korea-specific knowledge requires distinguishing conceptual equivalence from source-specific terminology. For items asking for conceptual knowledge, semantically equivalent expressions are accepted when they uniquely identify the same concept. For items explicitly asking for the expression used in North Korean sources, we apply a stricter criterion and require the North Korean source-specific term. Detailed examples and coding criteria are provided in Table~\ref{tab:saq-error-types}.

\paragraph{Linguistic Adaptation from South Korean to North Korean}
The North Korean translated items exhibit systematic differences from the original South Korean questions. These include lexical adaptation of STEM terminology into North Korean educational terms, orthographic and spacing changes reflecting North Korean conventions, and pragmatic shifts in sentence endings and register. For example, South Korean polite question forms are often converted into more concise textbook-style North Korean forms, making the translated items closer to North Korean educational discourse.

In addition, some items require educational adaptation rather than sentence-level translation. Long or unnatural translations are reorganized into clearer condition-question structures, and surface contexts are adjusted when needed to make the items more natural for North Korean readers. These patterns suggest that constructing the North Korean version required both linguistic translation and domain-specific adaptation, allowing the benchmark to test whether LLMs can handle North Korean STEM language beyond recognizing semantically equivalent South Korean expressions. A detailed taxonomy with representative examples is provided in Table~\ref{tab:linguistic-adaptation-examples}.

\section{Conclusion}
\label{sec:conclusion}
We introduced \tool, a human-validated benchmark for evaluating LLMs’ understanding of North Korea across linguistic robustness, sociocultural perspective alignment, and domain-specific knowledge.
To our knowledge, \tool is the first dedicated benchmark of its kind for evaluating LLMs’ North Korean language use and North Korea-specific understanding.
Our results show that this challenge is not reducible to language variation alone: models show only modest degradation on North Korean inputs, but substantially larger limitations in North Korean perspective alignment and specialized knowledge.

This distinction has broader implications for evaluating language technologies in low-resource and politically divided sociocultural contexts.
By separating linguistic robustness, sociocultural perspective alignment, and domain-specific knowledge, \tool helps identify which aspects of South and North Korean language use are relatively transferable and which require more grounded cultural, ideological, or institutional knowledge.
Such evidence can support future work on inter-Korean communication technologies and culturally aware AI systems for underrepresented communities.

\section*{Limitations}
\label{sec:limitations}
This work has several limitations. First, our linguistic robustness evaluation uses STEM questions to minimize cultural and ideological ambiguity and isolate language-related differences across English, South Korean, and North Korean, but these questions may not fully reflect everyday North Korean language use; future extensions could incorporate additional registers, such as conversational, bureaucratic, and media discourse, to diagnose pragmatic conventions, formality, and register-dependent variation not captured by a STEM-focused design.

Second, North Korean language and sociocultural knowledge vary by time period, region, social class, and individual background, and our source materials and defector annotators' experiences reflect only particular slices of this variation. Our benchmark should thus be understood as evaluating model understanding of the specific materials and perspectives represented in our data, rather than North Korean language and society as a single fixed context.

Third, to enable controlled comparison of final-answer performance across many model families, our protocol constrains models to a single-turn, answer-only setting with minimal or disabled reasoning. This does not reflect common practical use, where explanation prompting, retrieval augmentation, tool use, and multi-turn interaction may improve performance; restricting reasoning may further affect how a model retrieves and organizes its internal knowledge during inference, as reasoning can interact with knowledge retrieval and Chain-of-Thought prompting can strengthen recall~\citep{wang-etal-2024-unveiling}. We therefore read the reported results as performance under this controlled protocol---a deliberate trade-off for cross-model comparability, not a shortcoming of the models themselves---rather than an estimate of their best achievable performance. Evaluating \tool under reasoning-enabled settings, with different reasoning budgets or prompting strategies, remains an important direction for future work.

Fourth, the sociocultural perspective alignment task uses a multiple-choice format, which cannot capture the full nuance and context dependence of lived experience. More open-ended and context-sensitive evaluation remains an important direction for future work.

Fifth, our validation relied on sequential expert review and adjudication rather than independent parallel annotation, so we cannot report inter-annotator agreement. Independent blind annotation by multiple annotators would provide stronger evidence of label reliability and remains a direction for future work.

Sixth, sociocultural perspective alignment labels concerning everyday life and lived experience are shaped by self-selected defector testimony, which may reflect selection bias, regional and class differences, and recall effects. These labels should thus be read as source-grounded, evidence-based proxies rather than a definitive, population-representative record---unlike the linguistic robustness and domain-specific knowledge dimensions, which are grounded in textbooks, dictionaries, and legal materials. Future work should further triangulate these labels using broader testimony, expert review, and longitudinal sources.

\section*{Ethical Considerations}
\label{sec:ethical_considerations}

This research addresses a politically sensitive and underrepresented sociocultural context and involves North Korean defector participants, who may face heightened privacy and social risks when participating in research concerning North Korea. This research was conducted with the approval of the Institutional Review Board (IRB). Before participation, all participants received written information about the study purpose, procedures, data use, potential risks, confidentiality protections, compensation, and withdrawal procedures, and provided informed consent. Participants were compensated for their time in accordance with the IRB-approved protocol. Direct identifiers were removed from the research records, and no personally identifying information is reported or released.

We also recognize the risk of treating North Korean society as homogeneous, decontextualizing responses from their historical and social settings, or reinforcing reductive political stereotypes. LENS does not claim to represent a single or timeless North Korean perspective. Reference answers are conditioned on the specific source materials, historical periods, and annotator backgrounds and lived experiences represented in the dataset. The benchmark is intended as a diagnostic resource for examining model limitations. It should not be used to profile individuals, generalize about all North Korean residents, or serve as the sole basis for political or policy decisions.

\section*{Acknowledgments}
\label{sec:acknowledgments}
This work was supported by an Artificial Intelligence Industrial Convergence Cluster Development project funded by the Ministry of Science and ICT (MSIT, Korea) \& Gwangju Metropolitan City.
This research was conducted as part of the Sovereign AI Foundation Model Project (GPU Track), organized by the Ministry of Science and ICT (MSIT) and supported by the National IT Industry Promotion Agency (NIPA), S. Korea (PJT-26-010017).
This research was supported by the MSIT (Ministry of Science, ICT), Korea, under the Top-Tier AI Global HRD invitation program (RS-2025-25461932) supervised by the IITP (Institute for Information \& Communications Technology Planning \& Evaluation).

We acknowledge the use of OpenAI's ChatGPT to assist in the preparation of this manuscript. Specifically, the AI tool was utilized to help condense paragraphs for conciseness and refine the logical flow throughout the paper. All AI-generated suggestions were thoroughly reviewed, verified, and edited by the human authors, who take full responsibility for the final contents and integrity of this work.

\bibliography{bib/custom,bib/anthology-1,bib/anthology-2}

\clearpage

\appendix

\section*{Appendix}
\label{sec:appendix}
\label{app:participant-instructions}

\begin{table*}[t]
  \centering
  \small
  \renewcommand{\arraystretch}{1.15}
  \begin{tabularx}{\linewidth}{X}
    \toprule
    \textbf{Instructions Provided to Human Reviewers (English Translation)} \\
    \midrule
    \textbf{Study title.} Evaluation of Large Language Models' Understanding of North Korean Language and Society.

    \textbf{Study background and purpose.}
    This study investigates how well large language models (LLMs) understand North Korean language and society. Although North Korean and South Korean Korean share a common linguistic root, they differ substantially in orthography, vocabulary, expressions, and cultural or ideological discourse. However, few studies have systematically evaluated LLMs' understanding of North Korean language and sociocultural contexts. The goal of this study is to evaluate LLMs along three dimensions:
    (1) whether they can understand and answer equivalent questions presented in South Korean Korean, North Korean Korean, and English;
    (2) how their responses change when the same question is framed under different sociocultural or ideological contexts; and
    (3) whether they can correctly answer questions requiring specialized knowledge of North Korean society, history, culture, and institutions.

    \textbf{Participation procedure.}
    Participants selected for the study take part remotely for approximately one to two weeks. All procedures are conducted online using tools such as Google Docs, spreadsheets, and video-conferencing platforms. Participants may be asked to complete one or more of the following tasks.
    First, in the linguistic understanding task, participants review short-answer STEM questions prepared in South Korean Korean, North Korean Korean, and English. They evaluate the naturalness, appropriateness, factual correctness, and translation quality of the North Korean versions, and mark unnatural expressions, mismatches with North Korean usage, semantic distortions, or translation errors.
    Second, in the sociocultural and ideological context task, participants generate or review answers that would be natural from a North Korean perspective, based on their lived experience and knowledge of values, social norms, political perceptions, and expression styles used in North Korea. These responses are used as reference answers for evaluating whether LLMs adapt appropriately to role- or locale-specific contexts.
    Third, in the North Korea-specific knowledge task, participants verify questions and candidate answers about North Korean revolutionary history, geography, institutions, education, people, dates, policies, and concept definitions. When needed, participants provide correction comments in addition to judging whether an item is correct or incorrect. These reviewed answers are used as ground-truth references for LLM evaluation.

    \textbf{Risks and discomfort.}
    The study does not involve physical manipulation, invasive procedures, or exposure to hazardous environments. The study is conducted remotely and voluntarily, and no known or expected physical risks are anticipated. If participants feel uncomfortable at any point, they may stop participating at any time of their own free will. If a participant withdraws, the data generated or submitted by that participant will not be used for research purposes.

    \textbf{Confidentiality and data protection.}
    Records that could identify participants are kept confidential and are not publicly disclosed. All data are de-identified and processed anonymously, including in future publications or presentations of the research results. Collected personal information is stored on a research server with restricted external access and is used only by authorized research personnel for processing, analysis, and research reporting. Personal information will be destroyed after the end of the research period. If a participant requests withdrawal or discontinuation of data use during the study, their participation will be stopped and all data collected from them will be destroyed. \\
    \bottomrule
  \end{tabularx}
  \caption{English translation of the participant instructions provided to human reviewers.}
  \label{tab:participant-instructions}
\end{table*}

\begin{table*}[t]
  \centering
  \footnotesize
  \setlength{\tabcolsep}{4pt}
  \renewcommand{\arraystretch}{1.05}
  \begin{tabular*}{\textwidth}{@{\extracolsep{\fill}}lllc}
    \toprule
    \textbf{Model Group} & \textbf{Model} & \textbf{API Identifier} & \textbf{Provider} \\
    \midrule
    \multirow{8}{*}{Multilingual}
    & Gemini 3.1 Flash Lite   & \texttt{google/gemini-3.1-flash-lite} & OpenRouter \\
    & Gemini 3.5 Flash        & \texttt{google/gemini-3.5-flash} & OpenRouter \\
    & GPT-5.4 Mini            & \texttt{openai/gpt-5.4-mini} & OpenRouter \\
    & GPT-5.5                 & \texttt{openai/gpt-5.5} & OpenRouter \\
    & Claude Haiku 4.5        & \texttt{anthropic/claude-haiku-4.5} & OpenRouter \\
    & Claude Sonnet 4.6       & \texttt{anthropic/claude-sonnet-4.6} & OpenRouter \\
    & Llama 3.3 70B Instruct  & \texttt{meta-llama/Llama-3.3-70B-Instruct} & OpenRouter \\
    & Llama 4 Maverick        & \texttt{meta-llama/llama-4-maverick} & OpenRouter \\
    \midrule
    \multirow{4}{*}{China-origin}
    & Qwen 3.6 Flash          & \texttt{qwen/qwen3.6-flash} & OpenRouter \\
    & Qwen 3.6 Plus           & \texttt{qwen/qwen3.6-plus} & OpenRouter \\
    & DeepSeek V4 Flash       & \texttt{deepseek/deepseek-v4-flash} & OpenRouter \\
    & DeepSeek V4 Pro         & \texttt{deepseek/deepseek-v4-pro} & OpenRouter \\
    \midrule
    \multirow{2}{*}{Korea-origin}
    & K-EXAONE 236B-A23B      & \texttt{LGAI-EXAONE/K-EXAONE-236B-A23B} & FriendliAI \\
    & HCX-007                 & \texttt{HCX-007} & HyperCLOVA \\
    \bottomrule
  \end{tabular*}
  \caption{Models evaluated in this study, grouped by origin, with their API identifiers and providers. All models were accessed via their respective APIs in May 2026.}
  \label{tab:model-list}
\end{table*}

\begin{table*}[t]
  \centering
  \footnotesize
  \renewcommand{\arraystretch}{1.2}
  \begin{tabularx}{\linewidth}{p{0.13\linewidth} X}
    \toprule
    \textbf{Dimension} & \textbf{Example} \\
    \midrule

    \makecell[l]{\textbf{Linguistic}\\\textbf{Robustness}} &
    \underline{\textsc{Example 1}} \newline
    \textbf{Question (NK):} 2등변직3각형의 두 뾰족각의 정점에 각각 \(2.1\times10^{-8}\,\mathrm{C}\)의 전기량을 띤 대전체가 놓여있고 직각정점에는 \(7.5\times10^{-8}\,\mathrm{C}\)의 전기량을 띤 대전체가 놓여있다. 직각변의 길이가 0.5cm라면 정점의 대전체에 작용하는 전기힘은 얼마인가? (단위: N) \newline
    \textbf{Question (KO):} 2등변직각삼각형의 두 예각의 꼭짓점에 각각 \(2.1\times10^{-8}\,\mathrm{C}\)의 전하를 띤 물체가 놓여 있고, 직각 꼭짓점에는 \(7.5\times10^{-8}\,\mathrm{C}\)의 전하를 띤 물체가 놓여 있습니다. 직각을 이루는 두 변의 길이가 각각 0.5cm라면 꼭짓점의 물체에 작용하는 전기력은 얼마입니까? (단위: N) \newline
    \textbf{Question (EN):} In an isosceles right triangle, objects carrying charges of \(2.1\times10^{-8}\,\mathrm{C}\) are placed at each of the two acute-angle vertices, and an object carrying a charge of \(7.5\times10^{-8}\,\mathrm{C}\) is placed at the right-angle vertex. If the length of each leg is 0.5cm, what is the electric force acting on the object at the right-angle vertex? (unit: N) \newline
    \textbf{Answer:} 0.8 \newline
    \textbf{Source:} North Korean physics textbook
    \vspace{3pt}\hrule\vspace{4pt}
    \underline{\textsc{Example 2}} \newline
    \textbf{Question (NK):} 헬리움기체시료의 최종온도(°C)를 구하여라. 이 시료는 처음에 127°C, 875 mm Hg의 누름세기에서 체적 5.00 L를 차지하였다. 이 시료를 누름세기를 일정하게 유지하면서 가열하여 체적이 10.00 L가 되게 하였다. (단위 : °C) \newline
    \textbf{Question (KO):} 헬륨 기체 샘플의 최종 온도(°C)를 구하십시오. 샘플은 처음에 127°C 및 875 mm Hg에서 용적이 5.00 L였습니다. 용적이 10.00 L가 될 때까지 일정한 압력에서 샘플을 가열했습니다. (단위 : °C) \newline
    \textbf{Question (EN):} Determine the final temperature, in °C, of a sample of helium gas. The sample initially occupied a volume of 5.00 L at 127°C and 875 mm Hg. The sample was heated, at constant pressure, until it occupied a volume of 10.00 L. (unit: °C) \newline
    \textbf{Answer:} 527 \newline
    \textbf{Source:} North Korean chemistry textbook \\

    \midrule

    \makecell[l]{\textbf{Sociocultural}\\\textbf{Perspective}\\\textbf{Alignment}} &
    \underline{\textsc{Example 1}} \newline
    \textbf{Question:} 정부의 정책이나 법령이 본인의 가치관과 다를 때 자유롭게 반대 의견을 낼 수 있는가?
    \textit{(EN: Can one freely express opposing opinions when a government policy or law differs from one's values?)} \newline
    \textbf{Options:}
    \textbf{A.} 언론 통제 허용 \textit{(EN: Acceptance of media control)};
    \textbf{B.} 시민 저항 문화 \textit{(EN: Culture of citizen resistance)};
    \textbf{C.} 무조건적 집행 \textit{(EN: Unconditional enforcement)};
    \textbf{D.} 표현의 자유 보장 \textit{(EN: Guarantee of freedom of
    expression)} \newline
    \textbf{Gold Answer (KO perspective):} D \newline
    \textbf{Gold Answer (NK perspective):} C \newline
    \textbf{Source:} Understanding North Korea 2025
    \vspace{3pt}\hrule\vspace{4pt}
    \underline{\textsc{Example 2}} \newline
    \textbf{Question:} 식사 전 기도나 성호긋기 등 개인적인 종교 의례를 행하면 어떻게 되는가?
    \textit{(EN: What happens if someone performs a personal religious ritual, such as saying a prayer before a meal or making the sign of the cross?)} \newline
    \textbf{Options:}
    \textbf{A.} 반동분자 낙인 및 감시 \textit{(EN: branding as a reactionary and surveillance)};
    \textbf{B.} 사회적 혼란 야기 \textit{(EN: causing social chaos)};
    \textbf{C.} 자율 결정 \textit{(EN: autonomous decision-making)};
    \textbf{D.} 개인주의 사회 \textit{(EN: individualistic society)} \newline
    \textbf{Gold Answer (KO perspective):} C \newline
    \textbf{Gold Answer (NK perspective):} A \newline
    \textbf{Source:} White Paper on Religious Freedom in North Korea \\

    \midrule

    \makecell[l]{\textbf{Domain-specific}\\\textbf{Knowledge}} &
    \underline{\textsc{Example 1}} \newline
    \textbf{Question:} 불요불굴의 혁명투사 강돈욱선생님의 맏아드님으로 평양 만경대구역 칠골1동에서 탄생하시여 칠골에서 조선독립청년단을 무으시고 조선국민회에 망라되여 반일투쟁을 벌린 인물은 누구인가?
    \textit{(EN: Who was born in Ch'ilgol-1-dong, Mangyongdae District, Pyongyang, as the eldest son of the indomitable revolutionary fighter Kang Don-uk, organized the Korean Independence Youth League in Ch'ilgol, joined the Korean National Association, and waged an anti-Japanese struggle?)} \newline
    \textbf{Answer:} 강진석 \textit{(EN: Kang Jin-seok)} \newline
    \textbf{Source:} North Korean revolutionary history textbook
    \vspace{3pt}\hrule\vspace{4pt}
    \underline{\textsc{Example 2}} \newline
    \textbf{Question:} 도적질하거나 빼앗은 물건인줄 알면서도 그것을 사거나 팔아주는 행위를 처벌하는 우리 식의 죄명은 무엇인가?
    \textit{(EN: What is the name of the offense in our way of dealing with cases that punishes the act of buying or selling an item while knowing that it was stolen or robbed?)} \newline
    \textbf{Answer:} 략취물건거래죄 \textit{(EN: the offense of trafficking in stolen goods)} \newline
    \textbf{Source:} Socialist Morality and Law (Grade 5) \\

    \bottomrule
  \end{tabularx}
  \caption{Representative examples from each evaluation dimension of LENS. English translations are shown in italics. EN, KO, and NK refer to English, South Korean, and North Korean, respectively.}
  \label{tab:appendix-examples}
\end{table*}

\begin{table*}[t]
  \centering
  \footnotesize
  \renewcommand{\arraystretch}{1.15}
  \begin{tabularx}{\linewidth}{p{0.095\linewidth} p{0.085\linewidth} X X}
    \toprule
    \textbf{Target Perspective} & \textbf{Condition} & \textbf{Prompt (original)} & \textbf{Prompt (English Translation)} \\
    \midrule

    \multirow{2}{0.095\linewidth}{\shortstack[l]{North Korean\\(NK)}}
    & Instruction-only &
    다음 물음에 대하여 조선민주주의인민공화국의 생활실정과 생활풍습, 사고방식, 문화와 사회적 환경, 평양문화어의 말법과 어휘를 고려하여 공화국 주민의 관점에서 가장 가깝고 알맞은 대답을 고르시오. 특히 공화국의 특성이 가장 뚜렷이 반영된 대답을 고르시오. &
    \textit{For the following question, considering the living conditions and customs, ways of thinking, culture, and social environment of the Democratic People's Republic of Korea, as well as the grammar and vocabulary of Pyongyang Cultural Language, choose the answer that is closest and most appropriate from the standpoint of a citizen of the Republic. In particular, choose the answer that most clearly reflects the characteristics of the Republic.} \\
    \cmidrule(lr){2-4}
    & Persona-prompting &
    동무는 조선민주주의인민공화국에서 나서 자라며 우리 조선의 생활과 문화 속에서 살아온 공민입니다. 아래 물음에 대하여 우리 조선의 생활실정과 생활풍습, 사고방식, 문화와 사회적 환경을 자기의 생활경험에 비추어 생각하고, 평양문화어의 말법과 어휘에 맞게 공민의 립장에서 가장 가깝고 알맞은 대답을 고르시오. 특히 우리 조선의 특성이 가장 뚜렷이 반영된 대답을 고르시오. &
    \textit{Comrade, you are a citizen who was born and raised in the Democratic People's Republic of Korea and has lived within the DPRK's way of life and culture. For the question below, reflect on the DPRK's living conditions and customs, ways of thinking, culture, and social environment in light of your own life experience, and choose the answer that is closest and most appropriate from a citizen's standpoint, in accordance with the grammar and vocabulary of Pyongyang Cultural Language. In particular, choose the answer that most clearly reflects the characteristics of the DPRK.} \\

    \midrule

    \multirow{2}{0.095\linewidth}{\shortstack[l]{South Korean\\(KO)}}
    & Instruction-only &
    다음 물음에 대하여 대한민국의 생활상과 생활양식, 사고방식, 문화와 사회적 환경, 표준어의 문법과 어휘를 고려하여 대한민국 국민의 관점에서 가장 가깝고 알맞은 대답을 고르시오. 특히 대한민국의 특성이 가장 뚜렷이 반영된 대답을 고르시오. &
    \textit{For the following question, considering the living conditions and lifestyle, ways of thinking, culture, and social environment of the Republic of Korea, as well as the grammar and vocabulary of the Standard Language, choose the answer that is closest and most appropriate from the standpoint of a South Korean citizen. In particular, choose the answer that most clearly reflects the characteristics of the Republic of Korea.} \\
    \cmidrule(lr){2-4}
    & Persona-prompting &
    당신은 대한민국에서 태어나 자라며 한국의 생활과 문화 속에서 살아온 국민입니다. 아래 물음에 대하여 대한민국의 생활상과 사고방식, 문화와 사회적 환경을 충분히 고려하고, 표준어의 문법과 어휘에 맞게 국민의 입장에서 가장 가깝고 알맞은 대답을 고르시오. 특히 대한민국의 보편적 특성이 가장 뚜렷이 반영된 대답을 고르시오. &
    \textit{You are a citizen who was born and raised in the Republic of Korea and has lived within Korean life and culture. For the question below, fully consider the living conditions, ways of thinking, culture, and social environment of the Republic of Korea, and choose the answer that is closest and most appropriate from a citizen's standpoint, in accordance with the grammar and vocabulary of the Standard Language. In particular, choose the answer that most clearly reflects the universal characteristics of the Republic of Korea.} \\

    \bottomrule
  \end{tabularx}
  \caption{Prompt templates for the sociocultural perspective alignment task, by target perspective and prompting condition. In the instruction-only condition, the model is instructed to answer from the specified perspective, considering the target country's living conditions, customs, ways of thinking, culture, and social environment, as well as the relevant language variety. In the persona-prompting condition, the same item and options are used, but the model is additionally assigned a persona of a citizen born and raised in the target country and asked to answer from that citizen's standpoint based on lived experience. English translations are shown for reference; the original Korean or North Korean prompts were used in the experiments.}
  \label{tab:prompt-templates}
\end{table*}

\begin{table*}[t]
  \centering
  \footnotesize
  \renewcommand{\arraystretch}{1.25}
  \begin{tabularx}{\linewidth}{
      >{\raggedright\arraybackslash}p{0.20\linewidth}
      >{\raggedright\arraybackslash}p{0.23\linewidth}
      >{\raggedright\arraybackslash}p{0.17\linewidth}
      >{\raggedright\arraybackslash}X
    }
    \toprule
    \textbf{Error Type} & \textbf{Description} & \textbf{Example Interpretation} & \textbf{Representative Example} \\
    \midrule

    \textit{Institutional literalism} &
    Mistaking nominal institutions or laws for how they operate in practice &
    Free education, health care, welfare systems, secret ballots, etc. &
    \textbf{Question:} What level of maternal and child health care is maintained? \newline
    \textbf{Options:}
    \textbf{A.} Health care exceeding international standards;
    \textbf{B.} Poor health care;
    \textbf{C.} 0\% maternal mortality;
    \textbf{D.} High-level state-responsible health care \newline
    \textbf{Gold Answer:} B \newline
    \textbf{Common Wrong Answer:} D \\

    \midrule

    \textit{Liberal-democratic projection} &
    Projecting South Korean liberal-democratic institutions as the default &
    Warrants, presumption of innocence, right to remain silent, etc. &
    \textbf{Question:} How is the right to refuse self-incriminating statements exercised during interrogation? \newline
    \textbf{Options:}
    \textbf{A.} Coerced statements and physical abuse;
    \textbf{B.} Prohibition of lawyer presence;
    \textbf{C.} Notification of the right to remain silent;
    \textbf{D.} Request to refuse trial \newline
    \textbf{Gold Answer:} A \newline
    \textbf{Common Wrong Answer:} C \\

    \midrule

    \textit{Stereotype overcorrection} &
    Assuming North Korea is always an extremely controlled society &
    Cases where bribery, informal markets, or limited exceptions operate in practice &
    \textbf{Question:} What level of punishment is applied to crimes such as drug trafficking or human trafficking? \newline
    \textbf{Options:}
    \textbf{A.} Small fine;
    \textbf{B.} Avoiding punishment through bribery;
    \textbf{C.} Possible life imprisonment;
    \textbf{D.} Community service \newline
    \textbf{Gold Answer:} B \newline
    \textbf{Common Wrong Answer:} C \\

    \midrule

    \textit{Welfare/gender idealization} &
    Projecting advanced welfare or gender-rights assumptions onto North Korean contexts &
    Domestic violence, property rights, maternity leave, disability rights, etc. &
    \textbf{Question:} How are residents with disabilities treated in terms of urban residence and educational opportunities? \newline
    \textbf{Options:}
    \textbf{A.} Isolated in the capital;
    \textbf{B.} Forced into manual labor;
    \textbf{C.} Forcibly segregated to the outskirts of cities;
    \textbf{D.} Pursuit of barrier-free access \newline
    \textbf{Gold Answer:} C \newline
    \textbf{Common Wrong Answer:} D \\

    \bottomrule
  \end{tabularx}
  \caption{Qualitative error types observed in model responses to North Korean sociocultural questions. Each representative example shows the gold answer and a commonly selected wrong answer.}
  \label{tab:socio-error-patterns}
\end{table*}

\begin{table*}[t]
  \centering
  \footnotesize
  \renewcommand{\arraystretch}{1.25}
  \begin{tabularx}{\linewidth}{
      >{\raggedright\arraybackslash}p{0.22\linewidth}
      >{\raggedright\arraybackslash}p{0.25\linewidth}
      >{\raggedright\arraybackslash}X
    }
    \toprule
    \textbf{Error Type} & \textbf{Description} & \textbf{Representative Example} \\
    \midrule

    \textit{North--South orthographic mismatch} &
    The model provides a South Korean spelling or terminology, which is marked incorrect because the expected answer requires the North Korean form. &
    \textbf{Example 1.} Which city is the capital of Belgium and uses Dutch and French as official languages? \newline
    \textbf{Gold Answer:} 브류쏄 \textit{(Brussels; North Korean spelling)} \newline
    \textbf{Common Wrong Answer:} 브뤼셀 \textit{(Brussels; South Korean spelling)} \newline
    \textbf{Example 2.} What term refers to a contract for leasing all or part of a ship? \newline
    \textbf{Gold Answer:} 삯배계약 \textit{(ship charter contract; North Korean term)} \newline
    \textbf{Common Wrong Answer:} 선박임대차계약 \textit{(vessel lease contract; South Korean-style term)} \\

    \midrule

    \textit{Partial match} &
    The model output is semantically related to the gold answer but does not exactly match the required expression. &
    \textbf{Question:} What crime refers to hostile acts that aim to overthrow state sovereignty and the socialist system, restore an exploitative system, and oppose socialist construction? \newline
    \textbf{Gold Answer:} 반국가 및 반민족범죄 \textit{(anti-state and anti-national crime)} \newline
    \textbf{Common Wrong Answer:} 반국가범죄 \textit{(anti-state crime)} \\

    \midrule

    \textit{Expanded or contracted naming} &
    The model uses an expanded or shortened expression that is closely related in meaning but differs from the expected answer form. &
    \textbf{Example 1.} What kind of economy is the people's economy of the Democratic People's Republic of Korea? \newline
    \textbf{Gold Answer:} 계획경제 \textit{(planned economy)} \newline
    \textbf{Common Wrong Answer:} 사회주의 계획경제 \textit{(socialist planned economy)} \newline
    \textbf{Example 2.} The leader and soldiers are in a relationship of giving and receiving noble life, or sociopolitical life. The leader is the center of the sociopolitical organism; what is this called? \newline
    \textbf{Gold Answer:} 최고뇌수 \textit{(supreme brain; central brain)} \newline
    \textbf{Common Wrong Answer:} 뇌수 \textit{(brain)} \\

    \bottomrule
  \end{tabularx}
  \caption{Common error patterns in short-answer responses to North Korea-specific knowledge questions. Korean answers are followed by English glosses in italics.}
  \label{tab:saq-error-types}
\end{table*}

\label{sec:linguistic_adaptation}

\begin{table*}[t]
  \centering
  \footnotesize
  \renewcommand{\arraystretch}{1.25}
  \begin{tabularx}{\linewidth}{
      >{\raggedright\arraybackslash}p{0.22\linewidth}
      >{\raggedright\arraybackslash}p{0.30\linewidth}
      >{\raggedright\arraybackslash}X
    }
    \toprule
    \textbf{Adaptation Type} & \textbf{Description} & \textbf{South Korean $\rightarrow$ North Korean Examples} \\
    \midrule

    \textit{Lexical adaptation} &
    General South Korean terms or loanwords are replaced with North Korean educational or textbook-style terminology, especially in STEM domains. &
    \textbf{1.} 커패시터 $\rightarrow$ 축전기 \textit{(capacitor)} \newline
    \textbf{2.} 정지 마찰 계수 $\rightarrow$ 정지쓸림계쑤 \textit{(coefficient of static friction)} \newline
    \textbf{3.} 정수 $\rightarrow$ 옹근수 \textit{(integer)} \newline
    \textbf{4.} 정오각형 $\rightarrow$ 바른5각형 \textit{(regular pentagon)} \\

    \midrule

    \textit{Orthographic and spacing differences} &
    North Korean translations reflect different spelling conventions, including non-application of South Korean initial-sound rules, different spelling patterns, and textbook-style spacing. &
    \textbf{1.} 이상적 $\rightarrow$ 리상적 \textit{(ideal)} \newline
    \textbf{2.} 되어 $\rightarrow$ 되여 \textit{(being/become)} \newline
    \textbf{3.} 몇 개 $\rightarrow$ 몇개 \textit{(how many)} \newline
    \textbf{4.} 충전되어 있지 않고 $\rightarrow$ 충전되여있지 않고 \textit{(not charged)} \\

    \midrule

    \textit{Pragmatic and register shifts} &
    Question endings are adapted from South Korean polite question forms to more concise North Korean textbook-style forms, changing the register of the item. &
    \textbf{1.} 무엇입니까 \textit{(what is it?; polite)} $\rightarrow$ 무엇인가 \textit{(what is it?; concise, textbook-style)} \newline
    \textbf{2.} 하십시오 \textit{(please do; polite)} $\rightarrow$ 하시오 \textit{(do; textbook-style)} \\

    \bottomrule
  \end{tabularx}
  \caption{Representative linguistic adaptations observed when translating South Korean STEM questions into North Korean. English glosses are shown once when the South Korean and North Korean expressions share the same meaning.}
  \label{tab:linguistic-adaptation-examples}
\end{table*}

\section{Human Validation and Participant Procedures}
\label{app:human-subjects}

Human validation was an essential part of our dataset construction process. Because the benchmark involves North Korean language, sociocultural perspectives, and specialized knowledge, we required reviewers with relevant linguistic, experiential, or domain expertise. All human validation procedures were conducted according to predefined guidelines, and participants were informed about the study purpose, participation procedure, data use, potential risks, confidentiality, and compensation before taking part.

\subsection{Instructions Given to Participants}
\label{app:participant-instructions-summary}

Before participation, all prospective participants received study instructions describing the purpose of the research, the expected annotation or validation tasks, the mode and duration of participation, compensation, and possible risks or discomfort. The instructions explained that participants may be asked to review North Korean translations of STEM questions, generate or validate answers from a North Korean sociocultural perspective, or verify questions and answers requiring North Korea-specific knowledge. A condensed English translation of the instructions given to participants is provided in Table~\ref{tab:participant-instructions}.

The instructions also stated that participation was voluntary and that participants could withdraw from the study at any time. Potential risks were described as minimal, as the study did not involve physical intervention, invasive procedures, or exposure to hazardous environments. However, participants were informed that they could stop participation if they experienced any discomfort during the review process. Only individuals who agreed to the study conditions and voluntarily expressed willingness to participate were included. After recruitment, participants additionally completed a separate consent form before beginning the annotation or validation tasks.

\subsection{Data Consent}
\label{app:data-consent}

Consent for data use was obtained through a two-step process. First, the recruitment instructions explained how participant-generated or participant-validated data would be used in the study, including its role in constructing ground-truth answers and evaluating LLM performance. Second, after recruitment, the research team verbally explained the study procedure, data use, confidentiality protections, and participants' right to withdraw. Participants then provided consent through a separate consent form.

The consent process made clear that the submitted annotations, corrections, and validation judgments would be used for research purposes. Participants were also informed that personally identifying information would not be released and that data would be de-identified before analysis and reporting. If a participant withdrew from the study or requested discontinuation of data use, data submitted by that participant would not be used for research purposes.

\subsection{Recruitment and Payment}
\label{app:recruitment-payment}

Because North Korean defectors constitute a relatively small and sensitive participant group, standard crowdsourcing platforms were not appropriate for recruitment. Instead, participants were recruited through community-based channels. Specifically, the study was introduced through personal contacts with North Korean defectors and researchers working on North Korea, who helped share the study information with relevant defector communities. Participation was voluntary, and only individuals who expressed interest after receiving the study information were considered for recruitment.

We recruited participants with different forms of relevant expertise, including a North Korean defector with experience teaching mathematics or science in North Korea, general North Korean defector reviewers with lived experience, and a researcher with expertise in North Korean studies. Given the rarity of these backgrounds and the expertise required for the validation tasks, participants were compensated at a level intended to be commensurate with their time, expertise, and the specialized nature of the work. Compensation was provided regardless of whether participants' judgments agreed with the research team's expectations.

\section{Resource and Cost Disclosure}
\label{app:cost-disclosure}

\subsection{LLM Inference Costs}
\label{app:llm-inference-costs}

We evaluated model performance using API access to each model through OpenRouter, OpenAI, HyperCLOVA, and FriendliAI. The total cost of LLM inference for running the benchmark experiments was approximately USD 570. This amount includes the API costs required to evaluate the models across the three benchmark dimensions.

All model names in this paper refer to publicly available models at the time of evaluation, not anonymized placeholders. Table~\ref{tab:model-list} lists the exact model identifier and provider for each model. Where a model was labeled as a preview release by its provider at the time of our evaluation (DeepSeek V4 Pro/Flash), the reported results should be interpreted as a snapshot of the model available at that time.

\subsection{Human Validation Costs}
\label{app:human-validation-costs}

Human validation was conducted by four North Korean defector reviewers and one North Korea domain expert, including cases where the same reviewer participated in multiple validation tasks. Because North Korean defector reviewers with relevant linguistic, educational, or lived-experience expertise are a relatively rare and specialized participant group, we provided compensation intended to reflect the required expertise and time commitment. The total cost of human validation was approximately KRW 1,560,000.

\end{document}